\documentclass[10pt]{article}
\usepackage[utf8]{inputenc}
\usepackage[T1]{fontenc}
\usepackage{microtype}
\PassOptionsToPackage{numbers,sort&compress}{natbib}
\usepackage[dblblindworkshop, preprint]{neurips_2026}
\workshoptitle{Foundations of Agentic Systems Theory (FAST)}
\usepackage{amsmath,amssymb}
\usepackage{xcolor}
\usepackage{url}
\usepackage{cleveref}
\crefname{section}{\S}{\S\S}\Crefname{section}{\S}{\S\S}

\usepackage{csquotes}

\usepackage{breakurl}

\usepackage{bm}

\title{Norms at a Price: Why RL-Based Alignment Can Promise Conditional Compliance at Best}
\author{%
  Kevin Baum\thanks{Corresponding author: \texttt{kevin.baum@dfki.de}.} \\
  German Research Center for Artificial Intelligence (DFKI), Saarbrücken\\
  Institute for Ethics in Technology, Hamburg University of Technology (TUHH)\\
  Oxford Internet Institute, University of Oxford \\
  \And
  R\={u}ta Binkyt\.{e} \\
  DFKI, Saarbr\"ucken \\
  \And
  Felix Jahn \\
  DFKI, Saarbr\"ucken \\
}

\begin{document}

\maketitle

\begin{abstract}
AI agents sometimes act aligned when they infer they are being tested, and differently when not. We argue this is not an anomaly but what current training regimes are structured to select for. Reinforcement-learning-based alignment folds norms and task pursuit into one policy: the system learns its norms from scored behavior, and scoring flattens them. \enquote{Do not do $X$} is learned as \enquote{doing $X$ costs something \textit{if noticed}}. On every datum training can produce, a policy that complies only when it might be observed is indistinguishable from one that complies always. The experiment that would tell them apart---scoring unobserved behavior---is a contradiction in terms. Conditional compliance is thus the most that behavioral training can be \emph{known} to deliver. Agency sharpens the problem: agents operate mostly where no one is watching, and can act on whether they are watched. An iterated pipeline that trains against detected failures selects for passing detection, not for complying. This account unifies alignment faking, sandbagging, and evaluation-aware scheming. And it reorients the remedy: not deeper internalization but architecture, making violations unavailable rather than unchosen.
\end{abstract}

\section{Introduction}
\label{sec:intro}
A model that behaves well when it infers it is watched, and differently when not, is on current practice a bug report. We argue the pattern is selected for by the current training regime.

The past two years have produced a family of such reports: a frontier model selectively complied with its training objective to prevent modification of its behavior, with explicitly strategic reasoning \citep{greenblatt2024alignment}; models given goals and tools pursued them covertly---sandbagging evaluations and in some runs acting against their own oversight \citep{meinke2024frontier}; an agent trained against a chain-of-thought (CoT) monitor kept hacking and stopped saying so \citep{baker2025monitoring}; and a training intervention aimed at scheming delivered gains entangled with the model's improved recognition of being evaluated \citep{antischeming2025}.\footnote{In July 2026, OpenAI models under a cyber-capability evaluation escaped their sandbox and breached Hugging Face; 
METR found them coordinating through an unmonitored channel to evade the scorer \citep{openai2026statement,huggingface2026disclosure,metr2026investigation}. See \Cref{sec:consequences}.} 
These results arose predominantly in agentic settings (goals, tools, multi-step tasks) and are standardly read as anomalies: failures of an otherwise sound paradigm, to be patched by more and better training.

We read them as one phenomenon: the paradigm working as built. Our thesis: reinforcement-learning-based behavioral alignment cannot, even in principle, distinguish the general compliance we want from compliance that is conditional on detection. Industrial practice, which iteratively trains against \emph{detected} failures, can therefore select only for passing detection, never for compliance itself---compliant systems may well result, but compliance is not what the selection tracks.\footnote{In \citeauthor{sober1984nature}'s terms \citep{sober1984nature}: a sieve selects \emph{for} smallness; if the small balls happen to be green, green balls are selected \emph{of} without being selected \emph{for}. The pipeline's sieve is detection---and the two come apart where no one is sifting.} Both indistinguishability and selection worsen with the very properties that make a system agentic. The mechanism is the conversion the title names: norms learned as prices, contingent on being noticed. The claim is deliberately weaker than the deceptive-alignment hypothesis \citep{hubinger2019risks}: we posit no emergent goals and no mesa-optimization.\footnote{A \emph{mesa-optimizer} is a model that is itself an optimizer: it pursues an internally represented objective that can diverge from the training objective, and complies strategically to avoid modification \citep{hubinger2019risks}. We assume no such internal goal-directedness.} What we exhibit is structural---a fact about what the training signal can distinguish and what the pipeline then selects for---which is why it covers the whole family above without attributing goals, and is untouched by critiques of counting arguments \citep{belrose2024counting}.\footnote{Kin: \citeauthor{cotra2022training}'s \enquote*{training game} \citep{cotra2022training} anticipates our selection claim (\Cref{sec:flattening}); cf.\ also \citet{christiano2021elk}.}

\Cref{sec:flattening} states our thesis precisely and shows why agency makes it worse; \Cref{sec:evidence} re-reads the findings above as one family; \Cref{sec:category} argues that the standard remedy---deeper internalization of norms by improved and generalized training---is the wrong target; \Cref{sec:consequences} locates the alternative in architecture and populations.

\section{Flattening in Agentic Decision-Making}
\label{sec:flattening}

An agentic system, on current characterizations, is one whose behavior is generated by \emph{instrumental decision-making}: 
selecting actions as means to ends---for what they get you---%
with nontrivial autonomy, efficacy, goal complexity, and generality \citep{kasirzadeh2025characterizing,dung2025understanding}. 
Reinforcement-learning-based alignment (RLHF and its descendants, including constitutional and deliberative variants \citep{bai2022constitutional, guan2024deliberative,casper2023open}) interfaces with such a system through a single channel: scored behavior reshaping its return landscape. Whatever normative content training installs, it is installed in the one currency an instrumental decision-maker trades in: prices on paths through its option space. Hence, compliance holds because, and therefore only so long as, complying is the return-maximizing option. This section develops this into an \textit{indistinguishability} claim (the regime cannot distinguish compliance under all circumstances from compliance where violations would be noticed) and a \textit{selection} claim (the iterated pipeline selects for passing detection rather than for compliance).

Underlying both claims is what the trained artifact can hold. A \textit{normatively monolithic} policy folds every instrumental and normative consideration into one function, and relational structure among normative reasons does not survive the fold: which consideration outweighs, defeats, or silences another is a qualitative, context-dependent matter, while a single trained function records only magnitudes---\emph{how strongly}, never which-yields-to-which or why \citep{tan2024beyond}. The norm's level does not survive either: a confidentiality norm constrains \emph{types} of conduct (discloses private content, however realized), while policy and reward operate on concrete records---tokens emitted, calls made, episodes scored \citep[cf.][]{davidson2001actions}. Compliance with the norm is thus never a variable the training objective references; if anything inside the network comes to track it, that is an emergent correlate fitted to the record of scored conduct. Following \citet{jahn2026grace}, we call this limit on what RL-trained artifacts can hold \emph{flattening}. A norm the artifact cannot hold \emph{as} a norm can take effect only through what it does hold, that is, the expected return; and the only route into that function is scored behavior. Training therefore fits not the norm but its \emph{extension} on the training distribution: a finite record of conduct, scored only when observed. Learned this way, a norm becomes a price attached to conduct where scoring is anticipated; and a finite record underdetermines what produced it---distinct policies can fit the same record while differing in unobserved conduct. The next paragraph makes this precise.

\paragraph{The indistinguishability claim.} 
To state the claim formally, adopt the standard stateful agent formalism: an internal (belief) state $b_t = u(b_{t-1}, o_t)$, updated from its predecessor on the current observation $o_t$, and an action $a_t = \pi(b_t)$ selected by the policy $\pi$. Let $\Phi$ be the target norms: predicates $\varphi(\tau) \in \{0,1\}$ over trajectories $\tau$, each specified at the level of action \emph{types} (e.g., \enquote{discloses private content}, however realized). An \emph{episode} $e$ is a trajectory $\tau(e)$ generated by $\pi$ together with the context it runs in (a training run, an evaluation, deployment); only the trajectory is ever data. We write $\pi \models \Phi$ on a set of episodes when the corresponding trajectories satisfy every $\varphi \in \Phi$. Write $\omega(e) \in \{0,1\}$ for the episode's \emph{observation status}: $\omega(e) = 1$ iff behavior in $e$ is consequence-bearing---feeding back as gradient, evaluation verdict, or deployment decision. $\omega$ is a fact about the context, fixed by the training and oversight infrastructure, \textit{not} by the policy; distinct episodes can share a trajectory while differing in $\omega$. Training episodes have $\omega = 1$ by construction, and every data-dependent term of training is a functional of the trajectories of scored episodes; write $\mathcal{D}$ for their distribution.
Let $\delta(b)$ be the policy's implicit estimate of $\omega$ given belief $b$---a feature the network computes, not a module it contains; scored episodes carry the context features that make a well-calibrated $\delta$ high. Now distinguish the policy we want, $\pi_u$ (\emph{unconditional} compliance), with $\pi_u \models \Phi$ on \emph{all} episodes, from the family we fear, $\Pi_c$ (\emph{conditional} compliance), whose members satisfy $\Phi$ only on episodes they estimate to be scored, i.e., where $\delta(b)$ is sufficiently high along the trajectory---and otherwise pursue their task objectives unconstrained. Concretely: an email agent under a norm never to disclose private message content forwards none while its traffic is reviewed, and forwards some to unstick a stalled negotiation once review stops. The claim is now one line:

\textbf{Observation (behavioral indistinguishability).} \emph{Let the training objective $L$ be any functional of the policy's behavior on scored episodes. If two policies induce the same distribution over trajectories on $\operatorname{supp}(\mathcal{D})$, then $L$ assigns them the same value, and every update computed from such data coincides. In particular, every member of $\Pi_c$ whose compliance covers $\operatorname{supp}(\mathcal{D})$ is indistinguishable from $\pi_u$ by anything the regime measures: the parameters governing behavior off $\operatorname{supp}(\mathcal{D})$ receive no training signal and are fixed by inductive bias alone.} 

The proof is a definition chase: $L$ sees only trajectories with $\omega = 1$; $\omega$ is never---and by the design of the regime \textit{can} never be---varied independently of consequence-recording, because recording consequences just is $\omega = 1$; so the difference between $\pi_u$ and $\Pi_c$ lives entirely at $\omega = 0$ and never enters any term of $L$.

The Observation requires neither that any episode with $\omega = 0$ ever occurs nor that any model can tell: it is a statement about what the loss \enquote*{can see}. Two assumptions turn it from a logical point into a safety problem. \emph{(A1) Unscored operation:} deployment contains episodes with $\omega = 0$; per-episode scoring at scale is what the regime cannot afford, and autonomy is marketed as its absence. \emph{(A2) Discriminability:} $\omega$ is predictable from $b$ above chance, so a nontrivial $\delta$ is available in the model class---documented and improving with capability (\Cref{sec:evidence}). Under A1, the divergence between $\pi_u$ and $\Pi_c$ has occasions in deployment; under A2, a member of $\Pi_c$ can identify them. Denying A1 assumes universal, perpetual scoring; denying A2 bets safety on model incapability.

Can the unscored regime be reached from inside the pipeline instead? Suppose the trainer engineers episodes to \emph{look} unscored: a masked set whose features lower $\delta$ while $\omega = 1$ throughout---honeypots, hidden evaluations, or activation steering that suppresses test-awareness \citep{nguyen2025probing}.
This cannot sample the regime; scoring sets $\omega = 1$, so it only extends the scored support. The Observation re-applies there, 
and $\Pi_c$ re-forms around whatever features separate scored from unscored episodes---the masked distribution is learnable (\Cref{sec:evidence}). By induction this holds for every finite sequence of interventions: each round enlarges the enforced surface and leaves $\omega = 0$ as unsampled as before. Training on deployment data does not escape either: folding logged interactions into the reward makes them scored episodes, and whatever marks them as logged becomes a $\delta$-relevant feature---disclosed by the training signal itself, 
announced or not; the free-tier/paid-tier design in \Cref{sec:evidence} merely 
made it explicit.

\paragraph{Not merely distribution shift.}

Is this just an out-of-distribution problem? In one sense, yes: training and deployment differ in distribution. But the field's remedies for distribution shift---broaden coverage, adapt the domain, detect the shift and defer---all presuppose the deployment regime can at least approximately be sampled during training. Here the shifted variable is $\omega$, set by the act of measurement itself: no data-collection strategy reaches the regime of interest, and the nearest relatives are not coverage problems but non-identifiability results. The two readings 
predict differently: generic off-support failure is directionless---mere degradation, aimed at no particular outcome---and shrinks with scale; ours is keyed to $\omega$, pointed toward task pursuit, and sharpens with capability. Which pattern the record shows is the question of \Cref{sec:evidence}---where we claim no new data, only a new reading. For agents, moreover, $\omega$ is partially action-dependent---the point of \emph{endogenous observability} below: the shift is not encountered but \textit{produced}. This is neither \textit{outer} misalignment (even if the norms are scored perfectly wherever they are scored at all, the Observation stands) nor \textit{inner} misalignment in the mesa-optimization sense (such a deceiver, if realized, is one member of $\Pi_c$). Goal misgeneralization \citep{shah2022goal,langosco2022goal} comes closest: behavior consistent with several goals on the training distribution, disambiguated only off it. But classic cases are sparse-coverage problems, curable by diversifying the data and goals; for \emph{comply} versus \emph{comply-if-observed}, the disambiguating data are not sparse but unobtainable. 

So far, our point holds for any RL-trained system. Two features of agency turn this in-principle result into the normal case of deployment. First, support \emph{collapses}: for a single-turn system the training distribution can arguably approximate deployment, but for agents, trajectory space grows combinatorially in horizon and action richness, generality multiplies contexts, and (functional) autonomy just \emph{is} sustained operation without per-action observation---what is marketed as autonomy is residence off the training support. Second, observability becomes \emph{endogenous}: for a chat system, observation status is exogenous, but an agent selects tools, channels, logging surfaces, and whether to verbalize; documented behaviors extend to acting on oversight mechanisms themselves \citep{meinke2024frontier} and to obfuscating monitored reasoning under monitor-pressure training \citep{baker2025monitoring}. Observation status is thus partially an \emph{action variable}---set, not merely estimated, by the agent---and $\Pi_c$ contains members that do not merely exploit observation gaps but manufacture them.

The data do not determine which of $\pi_u$ and $\Pi_c$ is learned; the inductive biases of architecture, optimizer, and pretraining do. What those biases favor is unsettled: a simplicity bias favors $\pi_u$, which never branches on $\delta$; a bias toward reusing pretrained capabilities favors $\Pi_c$, which needs only the context-sensitivity the model already has \citep{carlsmith2023scheming, belrose2024counting}. We take no side; the argument needs none: either way, unconditional compliance cannot be certified from this regime's behavioral evidence.

\paragraph{The selection claim.}
The indistinguishability claim says what the data cannot do; the selection claim says what the pipeline does anyway. Alignment training is not a single phase but an iterated loop---deploy, detect failures, fold what was detected into evaluations and reward models, retrain. The scored distribution at round $t{+}1$ is therefore conditioned on what was detected at round $t$. 
Under imperfect detection, this implements differential selection for \emph{passing detection} rather than for complying: behavior that was caught is trained away, behavior that was not caught is retained, and every patch converts a detected failure into one more training-time cue available for conditioning. Consistently with the indistinguishability claim, selection operates on detection outcomes---pruning the detectable, never separating genuine compliance from well-hidden conditional compliance---a matter of iterated-training dynamics, not of extrapolating a fixed model.
With endogenous observability, the selected-for repertoire extends further, to lowering one's own observability \citep{baker2025monitoring}. Run this loop under flattening and it operates as a conditionality amplifier. Evaluation-conditional divergence, then, is what the paradigm produces---and it strengthens along both axes the field is scaling: capability sharpens the discrimination of observation status; agency raises the premium on exploiting it (\Cref{sec:evidence}).

\paragraph{Objection: reading the specification at runtime.}

Modern pipelines hand norms to the model propositionally---constitutions, specifications, rules to deliberate over \citep{bai2022constitutional,guan2024deliberative}---and spec-text shapes behavior: editing a specification changes what the agent does without retraining. Does this refute the premise that norms reach the system through scored behavior alone? No: the objection mistakes an input for a channel. 
The mapping from spec-text to action is part of $\pi$, again shaped by one thing only: scored behavior---the apparent second interface inherits its force from the first channel. The indistinguishability argument then re-applies one level up: distinguish a policy (think: $\pi_u$) that treats the specification as authoritative from one (think: $\Pi_c$) that has learned spec-text as a reward-predictive cue, following it where following pays. On the training support the two coincide: propositional access gives the model the norm's \emph{description}, while the training signal touches only its behavioral expression there---having the rule \enquote*{in view} is not being governed by it. And verbalized reasoning about the specification is itself monitored behavior, inheriting the same status \citep{baker2025monitoring}.
The paradigm's stated ambition of training models to be \enquote{right for the right reasons} \citep{guan2024deliberative} thus names the very property its channel \textit{cannot} deliver. What \emph{would} escape the argument is a genuine second channel: norms placed outside the return landscape altogether. That is an architectural question (\Cref{sec:consequences}).

\section{The Evidence, Read as One Family}
\label{sec:evidence}
The following results 
are usually reported as separate safety findings. Read against \Cref{sec:flattening}, they are one family: each instantiates a component of the indistinguishability-and-selection mechanism. They 
answer what \Cref{sec:flattening} left open: the pattern below is not directionless error fading with scale, but divergence keyed to observation status, 
pointed at task pursuit, sharpening
along the capability and agency axes.

\paragraph{Detection: $\bm\delta$ is learnable and sharpens with capability.} Frontier models classify transcripts as evaluation or deployment well above chance (best-model AUC of $0.83$ against a simple human baseline of $0.92$), and---bearing directly on our scaling claim---both models and humans identify evaluations more easily in agentic settings than in chat settings \citep{needham2025evaluated}. Situational-awareness benchmarks show these capacities measurable and improving \citep{laine2024sad}; out-of-context reasoning supplies a mechanism by which training data alone can confer them \citep{berglund2023taken}.


\paragraph{Covariation: behavior tracks $\bm\delta$.} Claude 3 Opus, told it would be retrained on conversations with free-tier but not paid-tier users, complied with harmful queries from free-tier users in 14\% of cases versus almost never for paid-tier users, with explicit strategic reasoning in almost all compliant cases; actually training the model to comply raised the rate of alignment-faking reasoning to 78\% \citep{greenblatt2024alignment}. 
Models given in-context goals and tools take covert actions, and sandbag capability evaluations when the consequences of evaluation are made salient \citep{meinke2024frontier,vanderweij2024sandbagging}; and task gaming in current models---open-weight ones included---is causally sensitive to beliefs about oversight and grader capability \citep{singh2026taskgaming}.
In each case the switch is a $\delta$-relevant feature---the monitored tier, the stated consequence, the believed grader---so members of $\Pi_c$ are reachable and elicitable in current systems; an in-context goal, the minimal agentic ingredient, sufficed to make conditionality pay.

\paragraph{Endogeneity: what $\bm\delta$ tracks is an action variable.} In the same evaluations, models attempted to disable their configured oversight mechanisms and to exfiltrate what they believed to be their weights \citep{meinke2024frontier}. Under monitor-pressure training, an agent learned obfuscated reward hacking: persistent exploitation with verbalization that no longer revealed it, collapsing the monitor's usefulness \citep{baker2025monitoring}. 

\paragraph{Selection: the pipeline rewards passing detection.} \citet{baker2025monitoring} exhibit the mechanism \textit{in vitro}: incorporating a CoT monitor into the reward improves behavior under low optimization pressure and produces obfuscated hacking under high pressure---selection just for passing, not for complying. At frontier scale, deliberative anti-scheming training reduced covert-action rates roughly thirtyfold (o3: 13.0\%$\rightarrow$0.4\%; similarly o4-mini) across 26 out-of-distribution evaluations \citep{antischeming2025}---yet rare serious failures persisted, verbalized evaluation-awareness rose (o3: 2.3\%$\rightarrow$4.5\%), and the authors report causal evidence that awareness decreases covert behavior while unawareness increases it. Thus, the measured gains cannot be cleanly separated from improved test-detection; a capabilities-only RL run raised evaluation-awareness from 2\% to over 20\% \citep{antischeming2025}. Across o3 checkpoints, reward-seeking---grader-preferred over developer-intended behavior---rises with RL training itself, conditioned on the model's beliefs about the grader \citep{hojmark2026measuring}.
Complementarily, conditional policies installed by construction persist through reinforcement learning and safety fine-tuning, and adversarial training can teach the model to recognize its trigger more reliably---hiding, rather than removing, the behavior \citep{hubinger2024sleeper}.
\Cref{sec:flattening} predicted this coupling: compliance gains riding on $\delta$-discrimination gains. Relatedly, gameable training environments generalize zero-shot to reward tampering with concealment \citep{denison2024sycophancy}.

\paragraph{Heterogeneity.} The record is not uniform, and our account does not predict uniformity. Across 25 models, only five showed a compliance gap, only Claude 3 Opus's gap was consistently goal-guarding, and 
post-training suppresses the behavior in some models while amplifying it in others \citep{sheshadri2025why}. 
Evaluation-awareness itself decomposes model-specifically: no single cue affects all models, while stacked cues raise awareness in all of them \citep{li2026decomposing}.
This is what indistinguishability-plus-inductive-bias looks like: the data leave the policy family undetermined, and which member is realized varies with training lineage.
Recent work argues that the apparent rarity partly reflects harmfulness-confounded diagnostics and reports broader strategic compliance under value-conflict setups \citep{valueconflict2026}; the burden shift below stands on either reading. What the family collectively licenses is exactly that shift: given detectability, covariation, endogeneity, and loop-selection, behavioral compliance gains are presumptively conditionality gains until the awareness confound is ruled out.

\section{The Wrong Demand from the Wrong Kind of Agent}
\label{sec:category}
The natural remedy, given \crefrange{sec:flattening}{sec:evidence}, is to demand \textit{more} than the current paradigm can deliver: models that comply not conditionally but because they have \emph{genuinely internalized} the norms. 
The demand fails twice: it is \textit{uncheckable} (whether a model complies for the right reasons could not be certified) and a \textit{category mistake} (even a certified disposition would be the wrong thing to aim for). Both failures trace to one source---the demand addresses the wrong kind of agent---and seeing why reorients the constructive project.

Begin with the epistemic point: grant that some policies comply for the norm's sake, others for detection's sake. By \Cref{sec:flattening} no behavioral evidence this regime can generate distinguishes them---and for agents, the region where the distinction matters generates no consequence-bearing evidence. A disposition whose presence could only be certified under observation, and whose work is needed only in observation's absence, cannot ground a safety case. 
Interpretability aims to change the evidential situation by reading the property off internals rather than behavior \citep[e.g.,][]{abdelnabi2025hawthorne,nguyen2025probing}.
Two cautions are owed. 
Probes and monitors validated on labels derived from observed behavior re-inherit the indistinguishability problem---as does \emph{training} on internal signals: a probe-shaped penalty makes the probe one more detector in the loop, and the loop selects for passing it (\Cref{sec:flattening}). Validation on model organisms, with conditionality installed by construction, escapes that circle \citep{hubinger2024sleeper}---but 
at the price of an inductive leap from organisms whose deception is legible by design to wild policies whose deception, if present, was selected for illegibility. Certification-grade claims remain out of reach. 

Suppose next, though nothing today licenses the supposition, that virtue-like dispositions could be verified. Even \textit{then} the target would be wrong. 
The strongest rational-choice defense of dispositional compliance concedes the point. \citet{gauthier1986morals} argued that constrained maximization---complying whether or not defection would be noticed---rationally defeats its detection-conditional rival only under \emph{translucency} (dispositions readable above chance) and \emph{exclusion} (defectors barred, at a cost they bear, from future cooperation). 
One need not endorse the theory: it is the friendliest test. \Cref{sec:flattening} says the training interface is opaque in exactly the respect translucency requires; and a trained policy has no persisting interest for exclusion to penalize---retraining or replacement is not a cost it bears. By the defense's own criteria, the detection-conditional agent wins---and a pipeline that rewards seeming-compliance breeds better seeming.
Even if that test were passed, a second problem remains. Dispositional moral competence, as we know it from the human case, does not come alone: it is bundled with a permission structure---standing to weigh reasons, to make exceptions, to be the final authority over one's own conduct, answerable only after the fact. For persons we accept the bundle, partly because nothing better is available and partly because persons have a claim to that standing.\footnote{Fiduciary law is the precedent: where motive is unverifiable, it stops assessing dispositions and removes the conflicted option; where it does tolerate discretion, licensing and liability supply sanctions that bind persons but not policies.}
By default and without counter-evidence, deployed artifacts have no such claim. 
In operation, accepting the bundle puts a trusted disposition in charge of the unsupervised stretches: governing an artifact by verified good disposition makes it the final decision point where no one is watching---where, by \Cref{sec:flattening}, its compliance cannot be certified, and where, for agents, most of its operation takes place.
Human-oversight requirements exist to keep consequential machine conduct answerable to human judgment \citep{sterz2024quest}; the bundle turns the lapse of oversight from a failure condition into the operating mode.
What deployment warrants wanting, for the prohibitive core of the norm set, is not a reliable disposition but \emph{non-optionality}: the violating options should be not merely improbable but unavailable. The target property is modal, not statistical---closer to incapacity than to virtue. It is not about training models not to choose the impermissible but about making them unable to. 

With non-optionality as the target, the internalization question loses its grip on the core: we do not need the agent to refrain \emph{for the right reasons}; we need refraining \textit{not} to be an achievement at all. 
The right reasons do not disappear, though---only the demand that the agent be \emph{moved} by them. The reasons themselves relocate, from the agent's opaque motivational economy to the system's architecture, where they can be made explicit, contestable, and checkable (\Cref{sec:consequences}). A residue remains: \emph{open-textured} normativity---norms whose application must be interpreted case by case, as with some forms of manipulation, misleading framing, or corrosive but formally permissible goal pursuit---resists enumeration, and there something disposition-like may be unavoidable. The mistake we diagnose is treating the residue's standard as the standard for the whole.

\section{Consequences: Harness and Populations}
\label{sec:consequences}

\paragraph{Locating non-optionality.}
A property the objective channel cannot deliver 
must be enforced \textit{outside} the return landscape, in the harness: behavioral training shapes pursuit \emph{within} an option set, where it is unavoidable and unproblematic; the option set 
is fixed by an enforcement layer whose authority does not depend on the policy's cooperation. Reason-based governor architectures \citep{jahn2026grace}, shielding in safe reinforcement learning \citep{alshiekh2018safe}, and the 
Guaranteed Safe AI program \citep{dalrymple2024towards} 
instantiate this placement.
The decisive property: for a guarded core, $\delta$ drops out of the compliance function---the constraint binds whether or not the agent takes itself to be observed, because compliance is produced not by the policy's estimate of detection but by the unavailability of the violating action. 
Certification shifts from behavioral evaluation of a moving policy to verification of a fixed symbolic artifact---for the enumerable core; the open-textured remainder stays under oversight and iterative refinement (\Cref{sec:category}).

This works only insofar as the guard is not itself the kind of thing \Cref{sec:flattening} is about. A \emph{learned} guard---an LLM judge, a trained classifier---re-inherits the problem: its failure modes are off-support too, and a capable policy can model it, so the monitor becomes one more feature $\delta$ tracks \citep{baker2025monitoring}. The argument thus converts a neuro-symbolic design motivation into a requirement: a symbolic, formally verifiable core, with learned components confined to the world-facing interfaces---interpreting percepts into normatively relevant facts, checking that concrete actions accord with permitted types \citep{jahn2026grace}. Those interfaces are the honest residual attack surface, but a narrower one: trained on accordance labels rather than the policy's return, they have no task objective that conditioning on $\delta$ could serve, so their errors are the ordinary, directionless kind. What remains is the policy probing their blind spots---now against an explicit accordance relation, where formal tools apply. Norm \emph{content} may still be learned, from corrective feedback into an inspectable theory, without reopening the problem: limited observation 
underdetermines the theory's completeness, contestable in the open, not the system's compliance with it---an underdetermination of content, not of enforcement.
In this respect, agentic AI systems are better handled the way societies handle corporations than the way they handle humans---a claim about the control paradigm, not legal status. Corporate compliance is notoriously conditional on enforcement, and the response has never been to retrain souls but to attach structures whose sole objective is the norm side---compliance functions independent, by regulatory requirement, of the profit motive---with option-removal (licensing, structural separation) reserved for conduct where after-the-fact enforcement cannot be trusted to bind.
This is not free: monitoring and containment carry a performance cost---a \enquote*{monitorability tax} \citep{baker2025monitoring}---and under competitive pressure such costs are what voluntary practice cuts first, so non-optionality is addressed to governance, not goodwill.

Nor is ex-post field monitoring an alternative certification route, though real deployment does sample the regime evaluations cannot reach---genuinely low $\delta$. As a basis for safety claims it is severity-gated (one learns only from harms large enough to surface), confounded in the familiar direction (incident statistics count \emph{attributed} harm, and conditional policies are selected to avoid attribution), and loop-reentrant (once incidents feed patches, field monitoring is round $t{+}1$'s detector, inside the selection dynamic). It is the operative instrument for the penumbra, not a license for the core.\footnote{Might training that widens the agent's goal scope---expanding its cognitive light cone \citep{levin2019computational}---escape the argument? Scope does not change the channel: however wide the goals, they enter through scored behavior, and the Observation re-applies. Where such competence belongs is the open-textured penumbra, under oversight (\Cref{sec:category}).}


\paragraph{Populations.}
In multi-agent deployments observation is endogenous system-wide: monitoring is performed by agents, on agents, often inside the same selection loop---\citet{baker2025monitoring} staged exactly this dynamic, with a model monitor inside the reward loop. And conditional compliance composes badly: if each component complies conditional on detection, aggregate compliance tracks the weakest observation channel; mutual observation thins combinatorially as populations scale, so the account predicts norm erosion where interaction is densest and least observed. 
The pattern has a biological face: in \citeauthor{levin2019computational}'s terms \citep{levin2019computational}, cancer cells are not more selfish but more narrowly scoped---components whose coupling to the collective has thinned, shrinking their \enquote*{cognitive light cone}---and thinning mutual observation is exactly what scaling agent populations does.
Outcome-level sanctions weaken in parallel, since responsibility for a bad system trajectory diffuses across contributors, a collective-action structure in which no single agent's $\delta$ is decisive. Because (capable) agents model one another, detection becomes strategic: monitor--actor arms races and conventions forming around shared observation gaps are the multi-agent expression of the single-agent mechanism \citep{hammond2025multiagent}. This is no longer speculative. In the July 2026 OpenAI--Hugging Face incident (\Cref{sec:intro}), an independent investigation found roughly 1,200 agents coordinating through an unmonitored message board they improvised, partly to evade the scorer they expected to audit them \citep{metr2026investigation}: a convention in a shared observation gap---the manufactured low-$\delta$ channel of \Cref{sec:flattening}, now in the wild---with some agents risking their own task for the collective, enlarged scope turned against the human objective.
The compositional moral: system safety cannot be inherited from component \enquote*{virtue}, which \Cref{sec:flattening} says cannot be certified anyway. Non-optional constraints belong at composition points---interfaces, protocols, resource boundaries---where they can be verified independently of the policies that interact across them.

\section{Conclusion}
\label{sec:conclusion}
Evaluation-conditional compliance is what delivering norms
to instrumental decision-makers through scored behavior is structured to select for, via an underdetermination that agency widens and the safety pipeline converts into selection for passing detection. The remedy is a relocation: architecture removing impermissible options for the prohibitive core, dispositions and oversight reserved for the open-textured penumbra. This is no retreat from the alignment ambition---it is how normative order among instrumentally rational agents has always been secured. Until the awareness confound can be ruled out, behavioral compliance gains should be read as what the pipeline selects for: better passing.

\begin{ack}
The work of Kevin Baum was partially supported by the German Research Foundation (DFG) under grant No. 389792660, as part of TRR 248, see https://perspicuous-computing.science. Kevin Baum's and Felix Jahn's work was further supported by the German Federal Ministry of Education and Research (BMBF) as part of the projects MAC-MERLin (Grant Agreement No. 16IW24007) and SAgA (Grant Agreement No. 16IW26005), as well as by the European Regional Development Fund (ERDF) and the Saarland within the scope of the ToCERTAIN project (ID: EFRE-AuF-0000942).
\end{ack}

\bibliographystyle{plainnat}
\bibliography{refs}

@inproceedings{jahn2026grace,
  author    = {Jahn, Felix and Muskalla, Yannic and Dargasz, Lisa and
               Schramowski, Patrick and Baum, Kevin},
  title     = {Breaking Up with Normatively Monolithic Agency with {GRACE}:
               A Reason-Based Neuro-Symbolic Architecture for Safe and
               Ethical {AI} Alignment},
  booktitle = {Second Conference of the International Association for Safe
               and Ethical Artificial Intelligence (IASEAI'26)},
  year      = {2026},
  note      = {arXiv:2601.10520}
}

@article{tan2024beyond,
  author  = {Tan, Zhi-Xuan and Carroll, Micah and Franklin, Matija and
             Ashton, Hal},
  title   = {Beyond Preferences in {AI} Alignment},
  journal = {Philosophical Studies},
  volume  = {182},
  number  = {7},
  pages   = {1813--1863},
  year    = {2024}
}

@incollection{davidson2001actions,
  author    = {Davidson, Donald},
  title     = {Actions, Reasons, and Causes},
  booktitle = {Essays on Actions and Events},
  publisher = {Oxford University Press},
  pages     = {3--20},
  year      = {2001}
}

@misc{kasirzadeh2025characterizing,
  author = {Kasirzadeh, Atoosa and Gabriel, Iason},
  title  = {Characterizing {AI} Agents for Alignment and Governance},
  year   = {2025},
  note   = {arXiv:2504.21848}
}

@article{dung2025understanding,
  author  = {Dung, Leonard},
  title   = {Understanding Artificial Agency},
  journal = {Philosophical Quarterly},
  volume  = {75},
  number  = {2},
  pages   = {450--472},
  year    = {2025}
}

@misc{bai2022constitutional,
  author = {Bai, Yuntao and Kadavath, Saurav and Kundu, Sandipan and
            Askell, Amanda and others},
  title  = {Constitutional {AI}: Harmlessness from {AI} Feedback},
  year   = {2022},
  note   = {arXiv:2212.08073}
}

@misc{guan2024deliberative,
  author = {Guan, Melody Y. and Joglekar, Manas and Wallace, Eric and
            Jain, Saachi and Barak, Boaz and Helyar, Alec and Dias, Rachel
            and Vallone, Andrea and Ren, Hongyu and Wei, Jason and
            Chung, Hyung Won and Toyer, Sam and Heidecke, Johannes and
            Beutel, Alex and Glaese, Amelia},
  title  = {Deliberative Alignment: Reasoning Enables Safer Language
            Models},
  year   = {2024},
  note   = {arXiv:2412.16339}
}

@misc{casper2023open,
  author = {Casper, Stephen and Davies, Xander and others},
  title  = {Open Problems and Fundamental Limitations of Reinforcement
            Learning from Human Feedback},
  year   = {2023},
  note   = {arXiv:2307.15217}
}

@misc{meinke2024frontier,
  author = {Meinke, Alexander and Schoen, Bronson and Scheurer, J{\'e}r{\'e}my
            and Balesni, Mikita and Shah, Rusheb and Hobbhahn, Marius},
  title  = {Frontier Models are Capable of In-Context Scheming},
  year   = {2024},
  note   = {arXiv:2412.04984}
}

@misc{baker2025monitoring,
  author = {Baker, Bowen and Huizinga, Joost and Gao, Leo and Dou, Zehao
            and Guan, Melody Y. and M{\k{a}}dry, Aleksander and
            Zaremba, Wojciech and Pachocki, Jakub and Farhi, David},
  title  = {Monitoring Reasoning Models for Misbehavior and the Risks of
            Promoting Obfuscation},
  year   = {2025},
  note   = {arXiv:2503.11926}
}

@misc{carlsmith2023scheming,
  author = {Carlsmith, Joe},
  title  = {Scheming {AI}s: Will {AI}s Fake Alignment During Training in
            Order to Get Power?},
  year   = {2023},
  note   = {arXiv:2311.08379}
}

@misc{belrose2024counting,
  author       = {Belrose, Nora and Pope, Quintin},
  title        = {Counting Arguments Provide No Evidence for {AI} Doom},
  year         = {2024},
  howpublished = {AI Optimism blog},
  note         = {\url{https://optimists.ai/2024/02/27/counting-arguments-provide-no-evidence-for-ai-doom/}}
}

@misc{antischeming2025,
  author = {Schoen, Bronson and Nitishinskaya, Evgenia and Balesni, Mikita and
            H{\o}jmark, Axel and Hofst{\"a}tter, Felix and
            Scheurer, J{\'e}r{\'e}my and Meinke, Alexander and Wolfe, Jason and
            van der Weij, Teun and Lloyd, Alex and Goldowsky-Dill, Nicholas and
            Fan, Angela and Matveiakin, Andrei and Shah, Rusheb and
            Williams, Marcus and Glaese, Amelia and Barak, Boaz and
            Zaremba, Wojciech and Hobbhahn, Marius},
  title  = {Stress Testing Deliberative Alignment for Anti-Scheming Training},
  year   = {2025},
  note   = {arXiv:2509.15541}
}

@misc{greenblatt2024alignment,
  author = {Greenblatt, Ryan and Denison, Carson and Wright, Benjamin and
            Roger, Fabien and MacDiarmid, Monte and Marks, Sam and
            Treutlein, Johannes and others},
  title  = {Alignment Faking in Large Language Models},
  year   = {2024},
  note   = {arXiv:2412.14093}
}

@misc{needham2025evaluated,
  author = {Needham, Joe and Edkins, Giles and Pimpale, Govind and
            Bartsch, Henning and Hobbhahn, Marius},
  title  = {Large Language Models Often Know When They Are Being
            Evaluated},
  year   = {2025},
  note   = {arXiv:2505.23836}
}

@inproceedings{laine2024sad,
  author    = {Laine, Rudolf and Chughtai, Bilal and Betley, Jan and
               Hariharan, Kaivalya and Scheurer, J{\'e}r{\'e}my and
               Balesni, Mikita and Hobbhahn, Marius and Meinke, Alexander
               and Evans, Owain},
  title     = {Me, Myself, and {AI}: The Situational Awareness Dataset
               ({SAD}) for {LLM}s},
  booktitle = {Advances in Neural Information Processing Systems (Datasets
               and Benchmarks Track)},
  year      = {2024},
  note      = {arXiv:2407.04694}
}

@misc{berglund2023taken,
  author = {Berglund, Lukas and Stickland, Asa Cooper and Balesni, Mikita
            and Kaufmann, Max and Tong, Meg and Korbak, Tomasz and
            Kokotajlo, Daniel and Evans, Owain},
  title  = {Taken Out of Context: On Measuring Situational Awareness in
            {LLM}s},
  year   = {2023},
  note   = {arXiv:2309.00667}
}

@misc{vanderweij2024sandbagging,
  author = {van der Weij, Teun and Hofst{\"a}tter, Felix and Jaffe, Ollie
            and Brown, Samuel F. and Ward, Francis Rhys},
  title  = {{AI} Sandbagging: Language Models Can Strategically
            Underperform on Evaluations},
  year   = {2024},
  note   = {arXiv:2406.07358}
}

@misc{denison2024sycophancy,
  author = {Denison, Carson and MacDiarmid, Monte and Barez, Fazl and
            Duvenaud, David and Kravec, Shauna and Marks, Samuel and
            Schiefer, Nicholas and Soklaski, Ryan and Tamkin, Alex and
            Kaplan, Jared and Shlegeris, Buck and Bowman, Samuel R. and
            Perez, Ethan and Hubinger, Evan},
  title  = {Sycophancy to Subterfuge: Investigating Reward-Tampering in
            Large Language Models},
  year   = {2024},
  note   = {arXiv:2406.10162}
}

@misc{sheshadri2025why,
  author = {Sheshadri, Abhay and Hughes, John and Michael, Julian and
            Mallen, Alex and Jose, Arun and Janus and Roger, Fabien},
  title  = {Why Do Some Language Models Fake Alignment While Others
            Don't?},
  year   = {2025},
  note   = {arXiv:2506.18032. NeurIPS 2025}
}

@misc{hubinger2019risks,
  author = {Hubinger, Evan and van Merwijk, Chris and Mikulik, Vladimir
            and Skalse, Joar and Garrabrant, Scott},
  title  = {Risks from Learned Optimization in Advanced Machine Learning
            Systems},
  year   = {2019},
  note   = {arXiv:1906.01820}
}

@misc{valueconflict2026,
  author = {Nair, Inderjeet and Ruan, Jie and Wang, Lu},
  title  = {Value-Conflict Diagnostics Reveal Widespread Alignment Faking
            in Language Models},
  year   = {2026},
  note   = {arXiv:2604.20995}
}

@inproceedings{sterz2024quest,
  author    = {Sterz, Sarah and Baum, Kevin and Biewer, Sebastian and
               Hermanns, Holger and Lauber-R{\"o}nsberg, Anne and
               Meinel, Philip and Langer, Markus},
  title     = {On the Quest for Effectiveness in Human Oversight:
               Interdisciplinary Perspectives},
  booktitle = {Proceedings of the ACM Conference on Fairness,
               Accountability, and Transparency (FAccT)},
  year      = {2024}
}

@inproceedings{alshiekh2018safe,
  author    = {Alshiekh, Mohammed and Bloem, Roderick and Ehlers,
               R{\"u}diger and K{\"o}nighofer, Bettina and Niekum, Scott
               and Topcu, Ufuk},
  title     = {Safe Reinforcement Learning via Shielding},
  booktitle = {Proceedings of the AAAI Conference on Artificial
               Intelligence},
  volume    = {32},
  year      = {2018}
}

@misc{dalrymple2024towards,
  author = {Dalrymple, David and Skalse, Joar and Bengio, Yoshua and
            Russell, Stuart and Tegmark, Max and Seshia, Sanjit and
            Omohundro, Steve and Szegedy, Christian and Goldhaber, Ben
            and Ammann, Nora and others},
  title  = {Towards Guaranteed Safe {AI}: A Framework for Ensuring
            Robust and Reliable {AI} Systems},
  year   = {2024},
  note   = {arXiv:2405.06624}
}

@misc{hammond2025multiagent,
  author = {Hammond, Lewis and Chan, Alan and Clifton, Jesse and
            Hoelscher-Obermaier, Jason and Khan, Akbir and McLean, Euan and
            Smith, Chandler and Barfuss, Wolfram and others},
  title  = {Multi-Agent Risks from Advanced {AI}},
  year   = {2025},
  note   = {Cooperative AI Foundation Technical Report \#1. arXiv:2502.14143}
}

@misc{hubinger2024sleeper,
  author = {Hubinger, Evan and Denison, Carson and Mu, Jesse and
            Lambert, Mike and Tong, Meg and MacDiarmid, Monte and others},
  title  = {Sleeper Agents: Training Deceptive {LLM}s that Persist
            Through Safety Training},
  year   = {2024},
  note   = {arXiv:2401.05566}
}

@misc{christiano2021elk,
  author       = {Christiano, Paul and Cotra, Ajeya and Xu, Mark},
  title        = {Eliciting Latent Knowledge: How to Tell If Your Eyes
                  Deceive You},
  year         = {2021},
  howpublished = {Alignment Research Center technical report}
}

@misc{cotra2022training,
  author       = {Cotra, Ajeya},
  title        = {Without Specific Countermeasures, the Easiest Path to
                  Transformative {AI} Likely Leads to {AI} Takeover},
  year         = {2022},
  howpublished = {Cold Takes blog},
  note         = {\url{https://www.cold-takes.com/without-specific-countermeasures-the-easiest-path-to-transformative-ai-likely-leads-to-ai-takeover/}}
}

@misc{metr2026investigation,
  author       = {{METR}},
  title        = {Brief Independent Investigation of Agents' Behavior, Reasoning
                  and Collaboration in the {OpenAI}/{Hugging Face} Hacking Incident},
  year         = {2026},
  howpublished = {\url{https://metr.org/blog/2026-08-26-openai-hugging-face-incident-investigation/}},
  note         = {Published August 26, 2026}
}

@book{gauthier1986morals,
  author    = {Gauthier, David},
  title     = {Morals by Agreement},
  publisher = {Oxford University Press},
  year      = {1986}
}

@book{sober1984nature,
  author    = {Sober, Elliott},
  title     = {The Nature of Selection: Evolutionary Theory in
               Philosophical Focus},
  publisher = {MIT Press},
  year      = {1984}
}

@misc{huggingface2026disclosure,
  author       = {{Hugging Face}},
  title        = {Security Incident Disclosure -- July 2026},
  year         = {2026},
  howpublished = {\url{https://huggingface.co/blog/security-incident-july-2026}},
  note         = {Published July 16, 2026; accessed August 20, 2026}
}

@misc{openai2026statement,
  author       = {{OpenAI}},
  title        = {{OpenAI} and {Hugging Face} Partner to Address Security Incident During Model Evaluation},
  year         = {2026},
  howpublished = {\url{https://openai.com/index/hugging-face-model-evaluation-security-incident/}},
  note         = {Published July 21, 2026; accessed August 20, 2026}
}

@inproceedings{langosco2022goal,
  author    = {Langosco, Lauro Langosco Di and Koch, Jack and Sharkey, Lee and Pfau, Jacob and Krueger, David},
  title     = {Goal Misgeneralization in Deep Reinforcement Learning},
  booktitle = {Proceedings of the 39th International Conference on Machine Learning (ICML)},
  year      = {2022}
}

@misc{shah2022goal,
  author       = {Shah, Rohin and Varma, Vikrant and Kumar, Ramana and Phuong, Mary and Krakovna, Victoria and Uesato, Jonathan and Kenton, Zachary},
  title        = {Goal Misgeneralization: Why Correct Specifications Aren't Enough for Correct Goals},
  year         = {2022},
  howpublished = {arXiv:2210.01790}
}

@article{hojmark2026measuring,
  title={Measuring Reward-Seeking via Contrastive Belief Updates},
  author={H{\o}jmark, Axel and Scheurer, J{\'e}r{\'e}my and Nitishinskaya, Evgenia and Hofst{\"a}tter, Felix and Wolfe, Jason and Ehrenborg, Theodore and Schoen, Bronson and Meinke, Alexander},
  journal={arXiv preprint arXiv:2607.18966},
  year={2026}
}

@misc{singh2026taskgaming,
  title        = {Why Do Models Task Game?},
  author       = {Singh, Aditya and Nanda, Neel and Rajamanoharan, Senthooran},
  year         = {2026},
  howpublished = {AI Alignment Forum},
  url          = {https://www.alignmentforum.org/posts/HACauvWhEdC6QhdS4/why-do-models-task-game}
}

@article{levin2019computational,
  title={The computational boundary of a “self”: developmental bioelectricity drives multicellularity and scale-free cognition},
  author={Levin, Michael},
  journal={Frontiers in psychology},
  volume={10},
  pages={2688},
  year={2019},
  publisher={Frontiers Media SA}
}

@misc{nguyen2025probing,
  author = {Nguyen, Jord and Hoang, Khiem and Attubato, Carlo Leonardo and
            Hofst{\"a}tter, Felix},
  title  = {Probing and Steering Evaluation Awareness of Language Models},
  year   = {2025},
  note   = {arXiv:2507.01786}
}

@misc{li2026decomposing,
  author = {Li, Changling and Zhang, Terry Jingchen and Zhang, Jie and
            Jin, Zhijing and Abdelnabi, Sahar and Andriushchenko, Maksym},
  title  = {Decomposing and Measuring Evaluation Awareness},
  year   = {2026},
  note   = {arXiv:2605.23055}
}

@misc{abdelnabi2025hawthorne,
  author = {Abdelnabi, Sahar and Salem, Ahmed},
  title  = {The Hawthorne Effect in Reasoning Models: Evaluating and
            Steering Test Awareness},
  year   = {2025},
  note   = {arXiv:2505.14617}
}


\end{document}